\documentclass[sn-basic, iicol]{sn-jnl}

\usepackage{graphicx}%
\usepackage{multirow}%
\usepackage{amsmath,amssymb,amsfonts}%
\usepackage{amsthm}%
\usepackage{mathrsfs}%
\usepackage[title]{appendix}%
\usepackage{textcomp}%
\usepackage{manyfoot}%
\usepackage{booktabs}%
\usepackage{algorithm}%
\usepackage{algpseudocode}%
\usepackage{listings}%
\usepackage{bm}
\usepackage{caption}
\usepackage{subcaption}
\usepackage[table]{xcolor}
\usepackage{colortbl}
\usepackage{balance}
\usepackage{color}
\makeatletter
\newcommand*\bigcdot{\mathpalette\bigcdot@{.5}}
\newcommand*\bigcdot@[2]{\mathbin{\vcenter{\hbox{\scalebox{#2}{$\m@th#1\bullet$}}}}}
\makeatother

\theoremstyle{thmstyleone}%
\theoremstyle{thmstyletwo}%

\theoremstyle{thmstylethree}%

\begin{document}

\title[Visual Content Refinement for CLIP Adaptation]{Rethinking Visual Content Refinement in  Low-Shot CLIP Adaptation}



\author[1]{\fnm{Jinda} \sur{Lu}}\email{lujd@mail.ustc.edu.cn}

\author*[1]{\fnm{Shuo} \sur{Wang}}\email{shuowang.edu@gmail.com}

\author[1]{\fnm{Yanbin} \sur{Hao}}\email{haoyanbin@hotmail.com}

\author[2]{\fnm{Haifeng} \sur{Liu}}\email{haifeng@leinao.ai}

\author[1]{\fnm{Xiang} \sur{Wang}}\email{xiangwang1223@gmail.com}

\author[3]{\fnm{Meng} \sur{Wang}}\email{eric.mengwang@gmail.com}

\affil[1]{\orgdiv{School of Cyber Science and Technology, School of Information Science and Technology, School of Artificial Intelligence and Data Science}, \orgname{University of Science and Technology of China}, \orgaddress{\street{No. 96, JinZhai Road}, \city{Hefei}, \postcode{230026}, \state{Anhui}, \country{China}}}

\affil[2]{\orgname{Brain-Inspired Technology Co., Ltd.}, \orgaddress{\street{No. 5089, Wangjiang West Road}, \city{Hefei}, \postcode{230088}, \state{Anhui}, \country{China}}}

\affil[3]{\orgdiv{School of Computer Science and Information Engineering}, \orgname{Hefei University of Technology}, \orgaddress{\street{ No. 485, Danxia Road}, \city{Hefei}, \postcode{230601}, \state{Anhui}, \country{China}}}

\abstract{Recent adaptations can boost the low-shot capability of Contrastive Vision-Language Pre-training (\textbf{CLIP}) by effectively facilitating knowledge transfer. However, these adaptation methods are usually operated on the global view of an input image, and thus biased perception of partial local details of the image. 
To solve this problem, we propose a \textbf{V}isual \textbf{C}ontent \textbf{R}efinement (VCR) before the adaptation calculation during the test stage. Specifically, we first decompose the test image into different scales to shift the feature extractor's attention to the details of the image. Then, we select the image view with the max prediction margin in each scale to filter out the noisy image views, where the prediction margins are calculated from the pre-trained CLIP model. Finally, we merge the content of the aforementioned selected image views based on their scales to construct a new robust representation. Thus, the merged content can be directly used to help the adapter focus on both global and local parts without any extra training parameters. 
We apply our method to 3 popular low-shot benchmark tasks with 13 datasets and achieve a significant improvement over state-of-the-art methods. For example, compared to the baseline (Tip-Adapter) on the few-shot classification task, our method achieves about 2\% average improvement for both training-free and training-need settings. Our code is available at: \url{https://github.com/injadlu/VCR}.}


\maketitle

\section{Introduction}\label{sec:intro}

\begin{figure*}[t]
\centering
\includegraphics[width=0.99\textwidth]{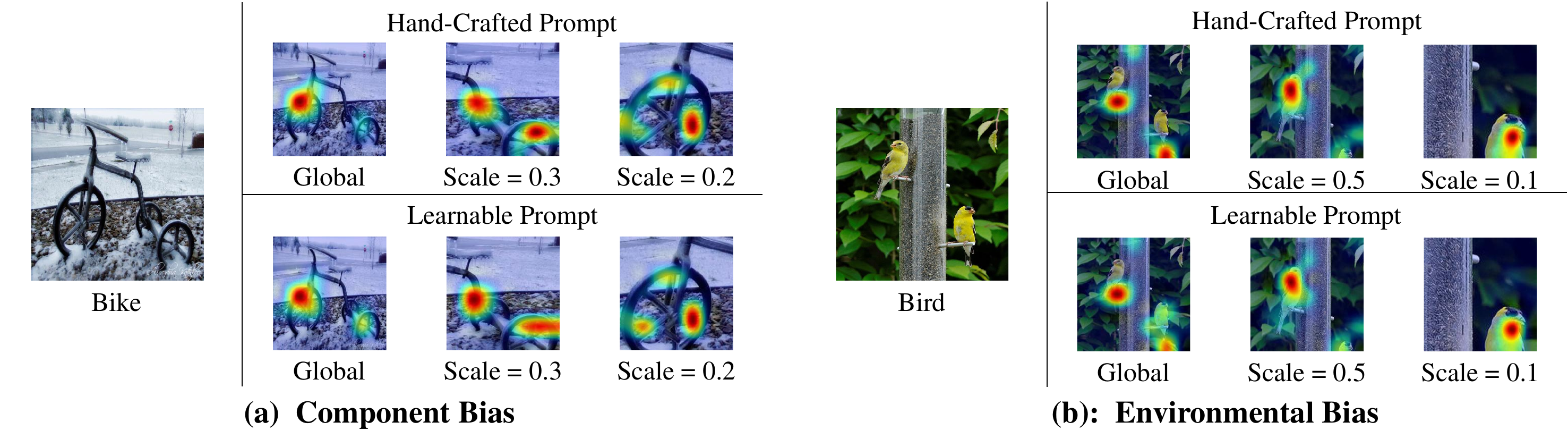}
\caption{
We use the responsive regions of images to visualize different perceived bias issues. Specifically, the responsive regions of the pre-trained CLIP model \cite{radford2021learning} are visualized by Grad-CAM \cite{selvaraju2017grad} from several samples from the validation set of ImageNet \cite{deng2009imagenet}, where \texttt{Component Bias} and \texttt{Environmental Bias} represent two perceived bias issues, \texttt{Hand-Crafted Prompt} \cite{zhang2022tip} and \texttt{Learnable Prompt} \cite{zhou2022learning} are two different prompt strategies. 
}
\label{fig: cam-result}
\end{figure*}

In low-shot tasks, previous methods such as \cite{lu2023semantic, chen2021meta, wang2023few} are faced restrictions with limited pre-training samples and categories, which brings restricted pre-trained knowledge. 
Recently, with the development of large-scale Vision-Language Models (VLMs), such as Contrastive Vision-Language Pre-training (CLIP) \cite{radford2021learning}, several methods \cite{zhang2022tip, zhou2022learning} aim to adapt CLIP for low-shot tasks and have achieved significant improvements.

In this paper, we systematically analyze current CLIP adaptation methods and discover a perceived bias issue in these adaptations.
For a better understanding, we first categorize them based on whether training is required. Concretely, one involves crafting prompt learners to effectively fine-tune the pre-trained CLIP model \cite{zhou2022learning}, while the other focuses on constructing training-free adapters with hand-crafted prompts \cite{zhang2022tip}. 
Then we visualize the responses on images of both prompt strategies by Grad-CAM \cite{selvaraju2017grad}, 
and find that this perceived bias issue can be manifested in the following aspects: 

\begin{itemize}
    \item \textbf{Component bias.} The adaptation model tends to concentrate on some small components of an object and overlooks the overall appearance of the object. As illustrated in Figure~\ref{fig: cam-result}(a, \texttt{Global}), the responsive regions are mostly distributed on the front fender while disregarding the overall structure of the Bike. This bias may mislead the prediction of the object, where the model inclines to predict objects with such components into the same category.
    \item \textbf{Environmental bias.} The adaptation model tends to prioritize environmental noises over the object. As shown in Figure~\ref{fig: cam-result}(b, \texttt{Global}), the model focuses on the branch rather than the bird itself (the responsive regions are mostly on the branch). This bias may influence the extraction of object features, where the model is prone to extract features containing such environmental noises.
\end{itemize}

\begin{table}[t]
\caption{Classification accuracy and distance comparisons between the global and multi-scale views, where each number denotes the mean of a category, and the categories are selected from the validation set of ImageNet \cite{deng2009imagenet}.}
\centering
\setlength{\tabcolsep}{1.2mm}{
\begin{tabular}{c | c | c c  c  c}
\hline\hline
\textbf{Category} & \textbf{Type} & \textit{\textbf{Acc}} & \textit{\textbf{Cos.}} & \textit{\textbf{L2}} & \textit{\textbf{KL}}\\ 
\hline
\multirow{2}*{\textbf{Bike}} & Global & 0.38 & 0.23 & 1.24 & 4.40 \\
& Multi-scale & \textbf{0.50} & \textbf{0.24} & \textbf{1.23} & \textbf{4.33}\\
\hline
\multirow{2}*{\textbf{Bird}} & Global & 0.88 & 0.26 & 1.19 & 1.37 \\
& Multi-scale & \textbf{0.90} & \textbf{0.29} & \textbf{1.21} & \textbf{1.13}\\
\hline
\multirow{2}*{\textbf{Giant Panda}} & Global & 0.92 & 0.26 & 1.19 & 0.80 \\
& Multi-scale & \textbf{0.96} & \textbf{0.28} & \textbf{1.21} & \textbf{0.50}\\
\hline
\multirow{2}*{\textbf{Ambulance}} & Global & 0.84 & 0.26 & 1.20 & 1.15 \\
& Multi-scale & \textbf{0.94} & \textbf{0.28} & \textbf{1.21} & \textbf{0.82}\\
\hline\hline
\end{tabular}
}
\label{tab:distance}
\end{table}

The aforementioned issue can be attributed to lacking comprehensive descriptions, which presents insufficient or excessive concentration on specific partial local details. To this end, we handcraft a multi-scale representation, combining image views from various local scales, and the comparisons between the \texttt{Multi-Scale} and \texttt{Global} representations are listed in Table~\ref{tab:distance}, where we compare both classification accuracy and distances between image and text features, including Cosine Distance (\textit{\textbf{Cos.}}), euclidean distance ( \textit{\textbf{L2}}), and KL Divergence, (\textit{\textbf{KL}}). We can observe that the multi-scale representation surpasses the global representation from all aspects, for instance, for the Bike category, \texttt{Multi-Scale} not just surpasses \texttt{Global} with 12\% classification accuracy, it also presents closer distances with the text feature from different aspects, which enhances the image representations.

Thus, we present a parameter-free approach to effectively introduce the multi-scale strategy in recent CLIP adaption methods, namely \textbf{V}isual \textbf{C}ontent \textbf{R}efinement (VCR). Specifically, our VCR consists of the following steps:

\begin{itemize}
    \item \textbf{Visual Decomposition (\ref{subsec:multi-scale construction}).} We decompose the input image into multiple scales, where the smaller scale retains more local details, while the larger scale contains more structural content. In Figure~\ref{fig: cam-result}(a), we can find that the larger local scale preserves the structural frame of the bike (Scale = 0.3), whereas the smaller local scale captures the detailed front wheel (Scale = 0.2). Both scales provide specific attributes of the bike that are neglected by the adaptation methods. Furthermore, to ensure stability for the subsequent content refinement process, we construct sufficient image views at each scale. We then refine the multi-scale content by carefully selecting the image view at each scale.
    \item \textbf{Content Refinement (\ref{subsec:features-select}).} During the refinement stage, we first utilize the pre-trained CLIP model to calculate the prediction scores of the image in each scale. Then, we apply the max-margin criterion to filter out noisy image views and retain the image view with the maximum prediction margin.
    For example, in Figure~\ref{fig: cam-result}(b), after the decomposition and careful refinement process, we discard the image views with the branch and precisely choose two birds at different scales (Scale = 0.1 or 0.5). This process shifts the adaptation model's focus on the bird object and eliminates the influences of environmental noises. Finally, we merge the visual features of the selected image views by combining them based on their scales to fuse different content of the input image. This operation enhances the stability of the input image, which preserves both the global structure and the local details, and thus helps the adaptation methods to achieve better recognition results.
\end{itemize}

The main contributions are threefold:
\begin{enumerate}
\item We decompose the input image into multiple scales to help the adaptation process alleviate the influences of the perceived bias issue in CLIP by discarding irrelevant noises and retaining more local details.

\item We design a refinement module to actively select the most relevant multi-scale content and merge different image views to refine the representation of the image. This further strengthens the adaptation process.

\item We conduct experiments on current CLIP adaptation methods, and our method achieves a significant improvement over current methods on three low-shot image recognition tasks with 13 benchmark datasets.
\end{enumerate}

\section{Related Works}
\label{sec:related}

\subsection{Vision-Language Models}
In recent years, Vision-Language Models (VLMs) have garnered significant attention due to their impressive performance \cite{radford2021learning, li2023blip, alayrac2022flamingo}. Notably, CLIP \cite{radford2021learning} and ALIGN \cite{jia2021scaling} have achieved remarkable zero-shot image recognition capabilities and strong transferability by aligning visual and textual embeddings in a contrastive manner using large-scale image-text pairs. Subsequent works \cite{mu2022slip, chen2023revisiting, li2022blip} have further improved the effectiveness of contrastive-based VLMs by enhancing CLIP's pre-training. For instance, the work in \cite{li2023scaling} introduces Mask Auto Encoder \cite{he2022masked} to the image encoder of CLIP, while SLIP \cite{mu2022slip} strengthens CLIP's pre-training with self-supervision loss. 
Building upon the success of Large-Language Models (LLMs) \cite{openai2023gpt4, brown2020language}, recent works \cite{zhu2023minigpt, instructblip} have begun to explore the use of LLMs to construct robust VLMs. 
Specifically, the works in \cite{alayrac2022flamingo, li2022blip} propose lightweight structures to align pre-trained LLMs with pre-trained visual encoders. Additionally, works such as \cite{instructblip, liu2024visual} construct powerful VLMs through instruction tuning \cite{wei2021finetuned}, where visual features are used as instructions to fine-tune LLMs.
Other studies \cite{menon2022visual, pratt2023does} generate informative category descriptions from LLMs to enhance the textual descriptions of contrastive-based VLMs.

\subsection{Adaptation for Vision-Language Models}
With the rise of Vision-Language Models (VLMs), numerous methods have emerged for adapting these models to various tasks \cite{patashnik2021styleclip, zhang2022pointclip, gao2022pyramidclip}. In this study, our focus is on the low-shot image recognition task, which has been explored through two different approaches.
The first approach involves prompt learners. CoOp \cite{zhou2022learning} is a seminal work in this area, where hand-crafted textual templates are replaced with continuous, learnable tokens. Based on CoOp, CoCoOp \cite{zhou2022conditional} proposes a lightweight network that generates image-conditional tokens. ProDA \cite{lu2022prompt} improves prompt learners by incorporating statistics, while works such as \cite{zhu2023prompt, khattak2023self, yao2023visual} address the overfitting phenomena in the prompt learning process.
The second approach focuses on training-free adapters. Tip-Adapter \cite{zhang2022tip} introduces a cache model that leverages a few training samples. This cache model is combined with the CLIP model for image recognition. TIP-X \cite{udandarao2023sus} enhances Tip-Adapter by replacing intra-modal distance with inter-modal distance. Another work \cite{zhu2023not} proposes a prior refinement module that selects discriminative feature channels to improve adaptation effectiveness. 
CaFo \cite{zhang2023prompt} further enhances adaptation capabilities through a cascade of foundation models.

\subsection{Multi-View Content Learning}
Our work draws inspiration from some multi-view content learning methods. Therefore, we list some related works in this subsection, such as \cite{luo2021rectifying, wang2020large, caron2020unsupervised}.
Specifically, the works in \cite{wang2020large, luo2021rectifying} split the foreground and the background views of an image to enhance the responsiveness to the foreground object, and the work in \cite{caron2020unsupervised} captures the relation between global and local views by contrastive training.

The differences between ours and these methods \cite{luo2021rectifying, wang2020large, caron2020unsupervised} are twofold: 
\textbf{(1)} They introduce training parameters and require additional computational costs, while ours is training-free and efficient.
\textbf{(2)} They focus on improving the effectiveness through backbone pre-training while ours achieves improvement through data refinement.
Moreover, compared to existing adaptation methods, which aim to fully utilize low-shot knowledge, we focus on addressing the perceived bias issue through decomposing the test samples.

\section{Approach}
\label{sec:method}

\begin{figure*}[t]
\centering
\includegraphics[width=0.96\textwidth]{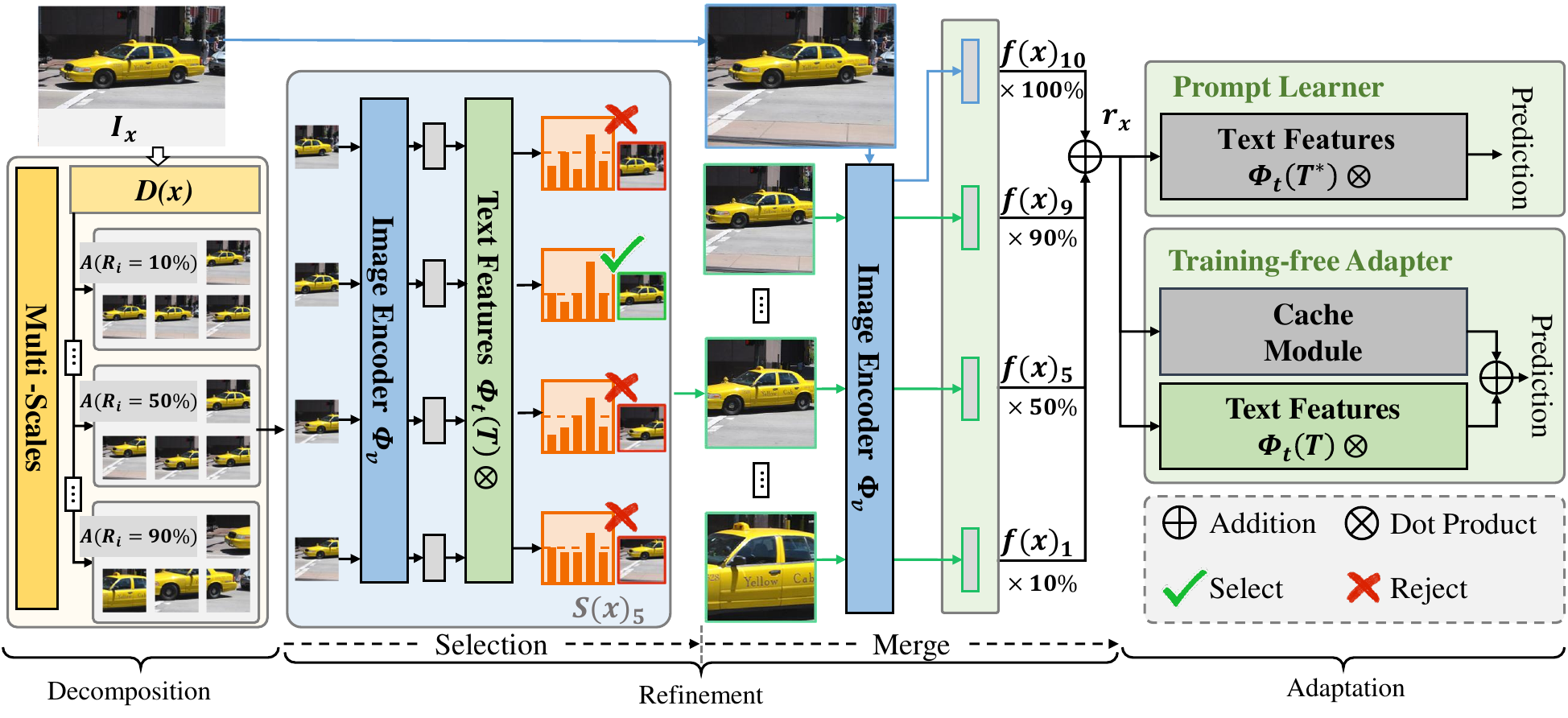}
\caption{An overview of our visual content refinement, given an image, we firstly decompose it into multiple scales, where each scale contains sufficient local views, then we refine the content at each scale, and finally we construct its refined representation to boost further adaptation methods.}
\label{fig: Framework}
\end{figure*}

In this section, we first overview our visual content refinement. Second, we revisit some preliminaries and then illustrate our method in detail. Finally, we describe the process of applying our method to existing methods.
As shown in Figure~\ref{fig: Framework}, given an image, we first decompose it into multi-scales, subsequently, we leverage the pre-trained CLIP model to select an image view with the max prediction margin in each scale, and finally, we merge the features of the selected image views with their scales to construct a new feature. We boost the adaptation models by replacing the original global feature with this carefully curated feature.

\subsection{Preliminary}
\label{subsec:preliminary}

\textbf{Contrastive Language-Image Pre-training (CLIP)} consists of an image encoder $\bm{\Phi}_{v}$ and a text encoder $\bm{\Phi}_{t}$.
After pre-training, CLIP can be effectively applied to downstream image recognition tasks with fixed encoders.
Specifically, given an image $\bm{I}_x$ and $C$ hand-crafted prompts $\bm{T}$ (\textit{e.g.} the word embedding of ``a photo of a $\left\{ class \right\}$'', where $\left\{ class \right\}$ is the category name), where $C$ is the number of categories, CLIP first represents them into $d$-dimensional features as $\bm{\Phi}_{v}(\bm{I}_x) \in \mathbb{R}^d $ and $\bm{\Phi}_{t}(\bm{T}) \in \mathbb{R}^d$. Then the CLIP classifier $\bm{\Gamma}_{\mathrm{CLIP}}(\bm{I}_x)$ can be formalized as:
\begin{eqnarray}
    \bm{\Gamma}_{\mathrm{CLIP}}(\bm{I}_x)  =  cos\langle\bm{\Phi}_{v}(\bm{I}_x),\bm{\Phi}_{t}(\bm{T})\rangle / \tau ,
\end{eqnarray}
where $\tau$ is the temperature coefficient learned in the pre-training phase, $cos \langle\bigcdot, \bigcdot\rangle$ denotes the cosine similarity, and the CLIP classifier $\bm{\Gamma}_{\mathrm{CLIP}}(\bm{I}_x)$ is utilized to classify $\bm{I}_x$ into $C$ different categories.

\subsection{Visual Decomposition}
\label{subsec:multi-scale construction}
To decompose the image into multi-scales, we first construct a decomposing scale set, which consists of $n$ different scales.
Specifically, the minimum decomposing scale of an image, denoted as $\alpha_{min}$, can be 1 pixel ($\alpha_{min} \to 0$), and the maximum decomposing scale, denoted as $\alpha_{max}$, infinitely approaching the entire image ($\alpha_{max} \to 1$). For convenient operation, the maximum scale $\alpha_{max}$ and the minimum scale $\alpha_{max}$ is set as 1 and 0, respectively. Then we divide the interval between the maximum and the minimum decomposing scale by $n$ for the partition ratio $\gamma$ calculation:
\begin{equation}
\label{equ:number n}
    \gamma = (\alpha_{max} - \alpha_{min}) / n .
\end{equation}
Extremely small scales of an image (\textit{e.g.} randomly sampled 1 pixel) can't provide useful content for the image, which potentially leads to misclassification of the object. Thus, we define the decomposing scale set ranging from $\alpha_{min} + \gamma$ to $\alpha_{max}$, which can be formalized as:
\begin{equation}
    \bm{\mathcal{R}} = \left \{ \alpha_{min}+\gamma; \alpha_{min}+2\gamma; \cdot\cdot\cdot;\alpha_{max} \right \} ,
\end{equation}
where $\bm{\mathcal{R}}$ contains $n-1$ local scales and $1$ global scale (Scale = 1). For convenient description, we use $i \ne n$ and $i = n$ to represent the local scales and the global scale, respectively.

Then, for the image $\bm{I}_x$, we decompose it into $n$ pre-defined scales with the decomposing scale set $\bm{\mathcal{R}}$.
Concretely, the decomposition process with pre-defined multiple scales can be calculated as:
\begin{eqnarray}
\label{equ: view construction}
\bm{D}(x)_{i} =
\left\{
\begin{aligned}
&\mathcal{A}(\bm{I}_x, \bm{\mathcal{R}}_i), & \text{if $i \ne n$}, \\ 
&\bm{I}_x, & \text{if $ i = n$}.
\end{aligned}
\right. 
\end{eqnarray}
Where $\mathcal{A}$ denotes a random cropping function, and it generates $m$ image views with the scale coefficient $\bm{\mathcal{R}}_i$ to restrict the crop size. $\bm{D}(x)$ is the multi-scale decomposed views of image $\bm{I}_x$, which contains $m$ image views for each local scale ($i \ne n$) and one image view for the global scale ($i = n$).
Thus the decomposition is achieved by the construction of $\bm{D}(x)$, and each scale in $\bm{D}(x)$ contains specific visual content. The decomposition process enables the model to attend to the content from different scales of the object and weakens the influences of irrelevant noises. 
Moreover, it contains sufficient views for each local scale, ensuring the stability of subsequent refinement operations. 

\subsection{Content Refinement}
\label{subsec:features-select}
Our content refinement consists of 2 parts: \textbf{(1)} content selection to filter out noisy image views, and \textbf{(2)} content merge to construct a robust visual feature. 
Our content selection is based on the max-margin criterion, therefore we first briefly illustrate the max-margin criterion. 
The max-margin criterion selects the image view with the largest margin between the highest and the secondary highest prediction probability \cite{tao2023boosting}. In essence, when the classification score between the highest category and the secondary highest category is significant, it indicates that the model exhibits a high level of confidence for the given sample, with a wider margin between the highest and other categories.
 
Then we elaborate on our content selection. 
For the image views of the $i$-th scale $\bm{D}(x)_i$ ($i\ne n$), we firstly employ the pre-trained CLIP model to calculate the classification logits as scores:
\begin{equation}
    \bm{S}(x)_i = \bm{\Gamma}_{\mathrm{CLIP}}(\bm{D}(x)_i),
\end{equation}
where $\bm{S}(x)_i$ contains $m$ classification logits to classify each image view into $C$ categories.
When the number of categories becomes large (\textit{e.g.} $C$ = 1000 in the ImageNet dataset \cite{deng2009imagenet}), the probability tends to converge toward uniformity, making the differences diminished with finite storage space, therefore we use the classification logits as scores.
Then for the $j$-th view in the $i$-th scale, the margin $\Delta S(x)_{i,j}$ can be formalized as:
\begin{equation}
\label{equ:margin calculation}
    \Delta  \bm{S}(x)_{i,j} = [\bm{S}(x)_{i,j}]_{top1} - [\bm{S}(x)_{i,j}]_{top2},
\end{equation}
where $[\bigcdot]_{top1}$ and $[\bigcdot]_{top2}$ are the highest score and the secondary highest score in $C$ categories, respectively. Thus, for the $i$-th scale ($i \ne n$), we select the image view with the max $\Delta  S(x)_{i}$ from $m$ image views.
For convenient illustration, we define this selected image view as the $k$-th image view as:
\begin{equation}
\label{equ:margin selection}
   k = argmax(\Delta  \bm{S}(x)_{i}).
\end{equation}

Finally, we merge the selected image views from different scales to construct a new robust representation.
Specifically, we first extract the visual feature of the selected image view at each scale. For the $i$-th scale ($i \ne n$), we denote the feature as $\bm{f}(x)_i = \bm{\Phi}_{v}(\bm{D}(x)_{i,k})$, and for $n$-th scale, we extract the visual feature as $\bm{f}(x)_n = \bm{\Phi}_{v}(\bm{I}_x)$.
Then we merge the visual features according to their corresponding decomposing scales to construct a refined feature $\bm{r}_x$ as:
\begin{equation}
\label{equ:merge}
     \bm{r}_x = \frac{\sum_{i=1}^{n} \bm{\mathcal{R}}_i \cdot \bm{f}(x)_i }{\sum_{i=1}^{n} \bm{\mathcal{R}}_i},
\end{equation}
where $\bm{\mathcal{R}}$ is the pre-defined decomposing scale set, and the refined feature $\bm{r}_x$ is utilized to substitute the original global one for adaptation methods.

\subsection{Application Process}
\label{subsec:adaptation}
In this subsection, we describe the application process of our refined features for training-free adapter \cite{zhang2022tip} and prompt learner \cite{zhou2022learning}. Specifically, the CLIP classifier with refined features can be formalized as:
\begin{eqnarray}
\label{equ:refined clip}
   \bm{\bar{\Gamma}}_{\mathrm{CLIP}}(\bm{I}_x)  =  cos\langle \bm{r}_x,\bm{\Phi}_{t}(\bm{T})\rangle / \tau.
\end{eqnarray}

For the training-free adapter, following \cite{zhang2022tip}, given the visual features and the one-hot labels of the training samples as $\bm{\mathrm{F}}_{\mathrm{train}}$ and  $\bm{\mathrm{L}}_{\mathrm{onehot}}$, respectively, the cache module with the refined features can be calculated as:
\begin{eqnarray}
    \bm{\Gamma}_{\mathrm{Cache}}(\bm{I}_x) = \varphi(cos\langle \bm{r}_x, \bm{\mathrm{F}}_{\mathrm{train}}\rangle)\cdot \bm{\mathrm{L}}_{\mathrm{onehot}},
\end{eqnarray}
where $\varphi( \cdot)$ is an exponential function.
Then, the training-free adapter can be calculated as the weighted addition of the cache module and the CLIP classifier by following \cite{zhang2022tip}. 

The prompt learner \cite{zhou2022learning} can be achieved by replacing the hand-craft prompts $\bm{T}$ with learnable tokens $\bm{T^{*}}$ in Equation~\ref{equ:refined clip}.
Compared to the original global feature, the multi-scale refined one enhances the representation of the objects by incorporating comprehensive local details while preserving global structural content, thus ensuring that the adaptation methods prioritize the object effectively.

\section{Experiments}
\label{sec:experiments}
In this section, we conduct experiments to validate the effectiveness of our \textbf{V}isual \textbf{C}ontent \textbf{R}efinement (VCR). Specifically, we first introduce the experimental settings, and next, we analyze the ablation studies, then we describe the comparison results with the state-of-the-art methods, and finally, we show the visualization results for detailed illustrations. 

\begin{table}[t]
\caption{Descriptions of the low-shot datasets.}
\centering
\setlength{\tabcolsep}{0.9mm}{
\begin{tabular}{l|c|c|c}
\hline\hline
\textbf{Datasets} & \textbf{Type} & \textbf{Categories} & \textbf{Test Data}\\
\hline
\textbf{Caltech-101}  & General & 100 & 2465\\
\textbf{DTD}  & Textures & 47 & 1692\\ 
\textbf{EuroSAT}  & Land & 10 & 8100\\
\textbf{FGVC-Aircraft}  & Aircraft & 100 & 3333\\
\textbf{Flowers102} & Flowers & 102 & 2463\\
\textbf{Food-101}  & Food & 101 & 30300\\
\textbf{ImageNet} & General & 1000 & 50000\\
\textbf{ImageNet-Sketch} & Sketch & 1000 & 50000\\
\textbf{ImageNetV2} & General & 1000 & 10000\\
\textbf{OxfordPets} & Pets & 37 & 3669\\
\textbf{StanfordCars} & Cars & 196 & 8041\\
\textbf{SUN397} & Scene & 397 & 19850\\
\textbf{UCF101} & Action & 101 & 3783\\
\hline\hline
\end{tabular}
}
\label{tab:datasets}
\end{table}

\subsection{Experimental Settings}
\label{subsec:exp_setting}
\textbf{Datasets.} To fully validate our method, we conduct experiments over 13 datasets spanning 3 benchmark tasks.
Specifically, we utilize 11 widely-used datasets for few-shot classification and base-to-new generalization, including 
ImageNet \cite{deng2009imagenet}, 
Caltech101 \cite{fei2004learning}, 
DTD \cite{cimpoi2014describing}, 
EuroSAT \cite{helber2019eurosat}, 
FGVCAircraft \cite{maji2013fine}, 
Food101 \cite{bossard2014food}, 
OxfordFlowers \cite{nilsback2008automated},
OxfordPets \cite{parkhi2012cats}, 
StanfordCars \cite{krause20133d}, 
SUN397 \cite{xiao2010sun}, 
UCF101 \cite{soomro2012ucf101}. 
And we employ ImageNet-Sketch \cite{wang2019learning} and ImageNetV2 \cite{recht2019imagenet} for cross-domain evaluation. Detailed descriptions of those low-shot datasets are listed in Table~\ref{tab:datasets}.

\noindent\textbf{Implementation Details.} 
We apply our method to current methods, therefore, all experimental settings and hyperparameters are based on the compared methods.
For visual decomposition, we set the number of scales $n$ as 10, and the number of image views $m$ as 100. We adopt the pre-trained CLIP model with the textual prompts in \cite{zhu2023not} for content refinement. 

\begin{table*}[t]
\centering
\caption{The accuracy (\%) of the classifiers evaluated with samples of different combination strategies.}
\label{tab:ab-all}
\scalebox{0.93}{
\begin{tabular}{l | c  c  c  c  c |  c  c  c  c  c}
\hline\hline
\multirow{2}*{\textbf{Methods}} & \multicolumn{5}{c|}{\textbf{Training-free}} & \multicolumn{5}{c}{\textbf{Training-need}}\\
 & 1 & 2 & 4 & 8 & 16 & 1 & 2 & 4 & 8 & 16 \\
\hline
Baseline & 60.70 & 60.96 & 60.99 & 61.44 & 62.02 & 61.12 & 61.71 & 62.57 & 63.87 & 65.46 \\
10-Crop & 61.51 & 61.91 & 61.92 & 62.26 & 62.92 & 62.21 & 62.74 & 63.52 & 64.88 & 66.50 \\
\hline
\rowcolor{lightgray}
\textbf{Decomposition} & 61.26 & 61.69 & 61.65 & 62.20 & 62.69 & 62.07 & 62.55 & 63.51 & 64.76 & 66.32 \\
\rowcolor{lightgray}
\textbf{Content Selection} & 63.17 & 63.37 & 63.48 & 63.78 & 64.07 & 63.47 & 63.81 & 64.20 & 65.33 & 66.65 \\
\rowcolor{lightgray}
\textbf{Content Merge} & \textbf{63.56} & \textbf{63.84} & \textbf{63.91} & \textbf{64.09} & \textbf{64.49} & \textbf{63.92} & \textbf{64.32} & \textbf{64.76} & \textbf{65.79} & \textbf{67.19} \\
\hline\hline
\end{tabular}
}
\end{table*}

\begin{table*}[t]
\caption{The accuracy (\%) of the classifiers evaluated with samples of different scales.}
\label{tab:ab1-FeaturePool}
\centering
\begin{tabular}{l | c  c  c  c  c |  c  c  c  c  c}
\hline\hline
\multirow{2}*{\textbf{Methods}} & \multicolumn{5}{c|}{\textbf{Training-free}} & \multicolumn{5}{c}{\textbf{Training-need}}\\
 & 1 & 2 & 4 & 8 & 16 & 1 & 2 & 4 & 8 & 16 \\
\hline
Scale=0.1 & 22.29 & 22.61 & 22.90 & 23.20 & 23.58 & 22.41 & 22.46 & 22.83 & 23.02 & 23.59 \\
Scale=0.2 & 38.61 & 39.02 & 39.25 & 39.79 & 40.36 & 38.93 & 39.16 & 39.77 & 40.36 & 41.55 \\
Scale=0.3 & 48.33 & 48.78 & 48.93 & 49.48 & 50.12 & 48.81 & 49.19 & 49.94 & 50.88 & 52.30 \\
Scale=0.4 & 53.67 & 54.15 & 54.26 & 54.81 & 55.37 & 54.29 & 54.70 & 55.52 & 56.61 & 58.14 \\
Scale=0.5 & 56.64 & 57.05 & 57.16 & 57.66 & 58.26 & 57.31 & 57.69 & 58.54 & 59.87 & 61.36 \\
Scale=0.6 & 58.70 & 59.05 & 59.09 & 59.65 & 60.17 & 59.38 & 59.78 & 60.62 & 61.90 & 63.40 \\
Scale=0.7 & 59.81 & 60.05 & 60.18 & 60.70 & 61.22 & 60.44 & 60.85 & 61.70 & 62.99 & 64.56 \\
Scale=0.8 & 60.03 & 60.37 & 60.39 & 60.95 & 61.39 & 60.68 & 61.13 & 61.97 & 63.20 & 64.77 \\
Scale=0.9 & 59.74 & 60.07 & 60.13 & 60.67 & 61.09 & 60.41 & 60.84 & 61.66 & 62.82 & 64.45 \\
Scale=1.0 & 60.70 & 60.96 & 60.99 & 61.44 & 62.02 & 61.12 & 61.71 & 62.57 & 63.87 & 65.46 \\
\hline
Multi-Crop & 60.93 & 61.43 & 61.29 & 61.87 & 62.33 & 61.82 & 62.09 & 63.21 & 64.45 & 66.00 \\
\rowcolor{lightgray}
\textbf{Multi-Scale} & \textbf{61.26} & \textbf{61.69} & \textbf{61.65} & \textbf{62.20} & \textbf{62.69} & \textbf{62.07} & \textbf{62.55} & \textbf{63.51} & \textbf{64.76} & \textbf{66.32} \\
\hline\hline
\end{tabular}
\end{table*}

\begin{table*}[t]
\caption{Experimental results with different number of $n$ (\%).}
\centering
\begin{tabular}{ l | c  c  c  c  c | c  c  c  c  c}
\hline\hline
\multirow{2}*{\textbf{Methods}} & \multicolumn{5}{c|}{\textbf{Training-free}} & \multicolumn{5}{c}{\textbf{Training-need}}\\
 & 1 & 2 & 4 & 8 & 16 & 1 & 2 & 4 & 8 & 16 \\
\hline
Baseline & 60.70 & 60.96 & 60.99 & 61.44 & 62.02 & 61.12 & 61.71 & 62.57 & 63.87 & 65.46 \\
$n$ = 5 & 63.37 & 63.52 & 63.65 & 63.98 & 64.42 & 63.68 & 64.08 & 64.55 & 65.64 & 67.01 \\
$n$ = 20 & \textbf{63.68} & \textbf{63.89} & 63.88 & \textbf{64.16} & \textbf{64.50} & 63.91 & \textbf{64.35} & \textbf{64.74} & \textbf{65.87} & 67.11 \\
\rowcolor{lightgray}
 $n$ = 10 & 63.56 & 63.84 & \textbf{63.91} & 64.09 & 64.49 & \textbf{63.92} & 64.33 & \textbf{64.74} & 65.79 & \textbf{67.19} \\
\hline\hline
\end{tabular}
\label{tab:n}
\end{table*}

\begin{table*}[t]
\caption{Experimental results with different selection criteria (\%).}
\centering
\begin{tabular}{l | c  c  c  c  c | c  c  c  c  c}
\hline\hline
\multirow{2}*{\textbf{Methods}} & \multicolumn{5}{c|}{\textbf{Training-free}} & \multicolumn{5}{c}{\textbf{Training-need}}\\
 & 1 & 2 & 4 & 8 & 16 & 1 & 2 & 4 & 8 & 16 \\
 \hline
Baseline & 60.70 & 60.96 & 60.99 & 61.44 & 62.02 & 61.12 & 61.71 & 62.57 & 63.87 & 65.46 \\
Min-Margin & 57.30 & 57.75 & 58.08 & 58.65 & 59.45 & 58.34 & 58.91 & 60.33 & 61.88 & 63.69 \\
Min-Entropy & 62.05 & 62.33 & 62.37 & 62.98 & 63.40 & 62.85 & 63.27 & 64.24 & 65.62 & 67.05 \\
\rowcolor{lightgray}
 Max-Margin & \textbf{63.56} & \textbf{63.84} & \textbf{63.91} & \textbf{64.09} & \textbf{64.49} & \textbf{63.92} & \textbf{64.33} & \textbf{64.74} & \textbf{65.79} & \textbf{67.19} \\
\hline\hline
\end{tabular}
\label{tab:select}
\end{table*}

\subsection{Ablation Studies}
\label{subsec:exp_ablation}
In this subsection, we use ImageNet to validate the effectiveness of different components of our method. Specifically, we conduct experiments under the few-shot classification benchmark with 1/2/4/8/16-shot training samples for evaluation, where the training set and the testing set are constructed by following \cite{zhang2022tip}. 

All experiments in the ablation studies are conducted by using Tip-Adapter \cite{zhang2022tip} with ResNet50 \cite{he2016deep} and the modified transformer \cite{radford2019language} as visual and textual encoder, respectively, under both training-free and training-need settings.

\subsubsection{Analysis of different components.}
\label{subsubsec:different components}
In this ablation study, we evaluate different components of our method, and the experimental results are listed in Table~\ref{tab:ab-all}. Specifically, ``Baseline'' shows the result with the global feature, ``Decomposition'' implies that using the average of one randomly selected feature from each scale, and ``Content Selection'' is constructed as the average of the selected features with the max-margin criterion, 
``Content Merge'' shows the results of the selected features with Equation~\ref{equ:merge}. Moreover, we also list the results with other crop-based evaluation strategies (\textit{e.g.} 10-crop \cite{he2016deep}) for a comprehensive comparison. 

From Table~\ref{tab:ab-all}, we can observe that our strategies bring steady performance improvements, it's worth noting that our selection yields significant enhancement, and it becomes more effective with limited samples (\textit{e.g.} 1/2/4 samples), which achieves the most performance improvement in for 1-shot with 1.9\% and 1.4\% gain in the training-free and training-need setting, respectively.
Moreover, compared with the ``10-crop'', combining all our strategies (``Content Merge'') yields an average of more than 2\% gains for all shots in the training-free setting and more than 1.5\% improvements for all shots in the training-need setting. 
These improvements are significant, and we believe that the reasons are twofold: \textbf{(1)} Our strategies align adaptation with pre-training, where the CLIP model pre-trained with 10-cropped image views to focus more on multi-scale content, and \textbf{(2)} Our strategies relieves the perceived bias issue, which eliminates the influences of the environmental noises, and helps the adaptation methods concentrate on different components of the objects.

\subsubsection{Effects of visual decomposition.}
\label{subsubsec:ab1 decomposition}
In this ablation, we validate the effects of decomposing input images into different scales. For each scale, we randomly select 1 image view from 100 image views by Equation~\ref{equ: view construction} and then extract its visual feature. 
``Multi-Scale'' denotes using an average of one randomly selected feature from each scale. 
Meanwhile, to avoid the influence of randomness, we conduct 10 experiments for all random selections and report the mean accuracy. 
For comparison, we also list the results of ``Multi-Crop'', where ``Multi-Crop'' utilizes the average feature of randomly cropped 100 image views from various scales. The experimental results are displayed in Table~\ref{tab:ab1-FeaturePool}, where ``Scale=1.0'' denotes the result with the global feature.

Specifically, we can find that with the increase of scale, the performance gradually improves, and as the ``Scale'' grows to 0.7, the performance stabilizes. The reason is that adaptation methods with small scales are more susceptible to component bias, small scales offer insufficient descriptions of the objects, and the overall global structure is neglected.
Meanwhile, we also observe that ``Multi-Scale'' achieves further improvement over ``Multi-Crop'', we suppose that this may be related to the uncontrollable scale in ``Multi-Crop''.
Furthermore, it's noteworthy that ``Multi-Scale''  requires a smaller number of image views compared to ``Multi-Crop'', the latter utilizes 100 image views, whereas the former is constructed with only 10 image views from different scales.

\subsubsection{Influences of content refinement.}
In this ablation, we validate the influences of content refinement from 2 aspects: \textbf{(1)} sensitive to various numbers of $n$, and \textbf{(2)} comparison with different selection criteria.

\noindent \textbf{Sensitive to various numbers of $n$.} 
In this experiment, we evaluate different numbers of $n$ for decomposing scale set construction by Equation~\ref{equ:number n}, where $n$ denotes the number of scales, and we select $n$ = 5/10/20, the experimental results are shown in Table~\ref{tab:n}. Moreover, for convenient description, we also show the results with the global feature, denoted as ``Baseline''. 

Specifically, we can find that compared with ``Baseline'', all numbers of $n$ achieve significant improvements, this not only demonstrates the effectiveness of our method, but also shows that our method can perform well without adjustment of hyper-parameter. Meanwhile, as $n$ increases to 10, the performance improves and stabilizes. Thus, we set $n$ to 10 for balancing accuracy and costs.

\noindent \textbf{Comparison with different selection criteria.}
In this experiment, we evaluate different strategies for image view selection. For a fair comparison, we utilize the Min-Entropy, Min-Margin, and Max-Margin criteria for selection, and we also list the ``Baseline'' result, which employs the global feature only, the results are illustrated in Table~\ref{tab:select}. 
Specifically, we can observe that \textbf{(1)} compared with ``Baseline'', the Min-Margin decreases the performance, which misleads the adaptation process and implies that the Min-Margin brings noises to the representation. And \textbf{(2)} Max-Margin achieves the best performance, which improves all other methods with a large margin. Compared with the entropy minimization criterion, the Max-Margin achieves more than 1\% gains, which fully demonstrates our effectiveness.

\begin{figure*}[t]
\centering
\includegraphics[width=0.98\textwidth]{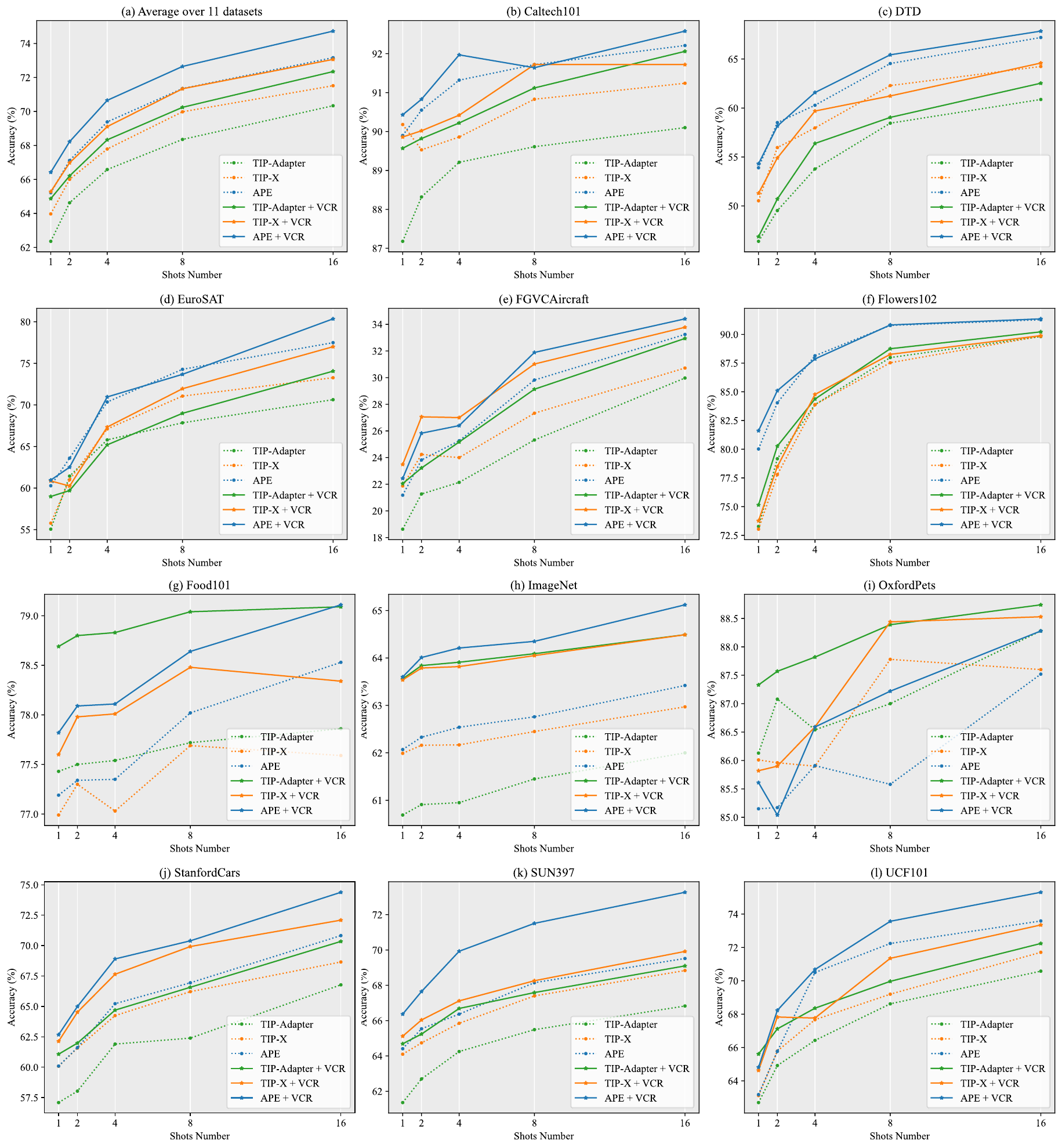}
\caption{Few-shot performance with training-free methods on 11 datasets, we first show the average results, and the following are organized in the order of dataset names, zoom in for clear recognition.}
\label{fig: training-free}
\end{figure*}

\begin{figure*}[t]
\centering
\includegraphics[width=0.98\textwidth]{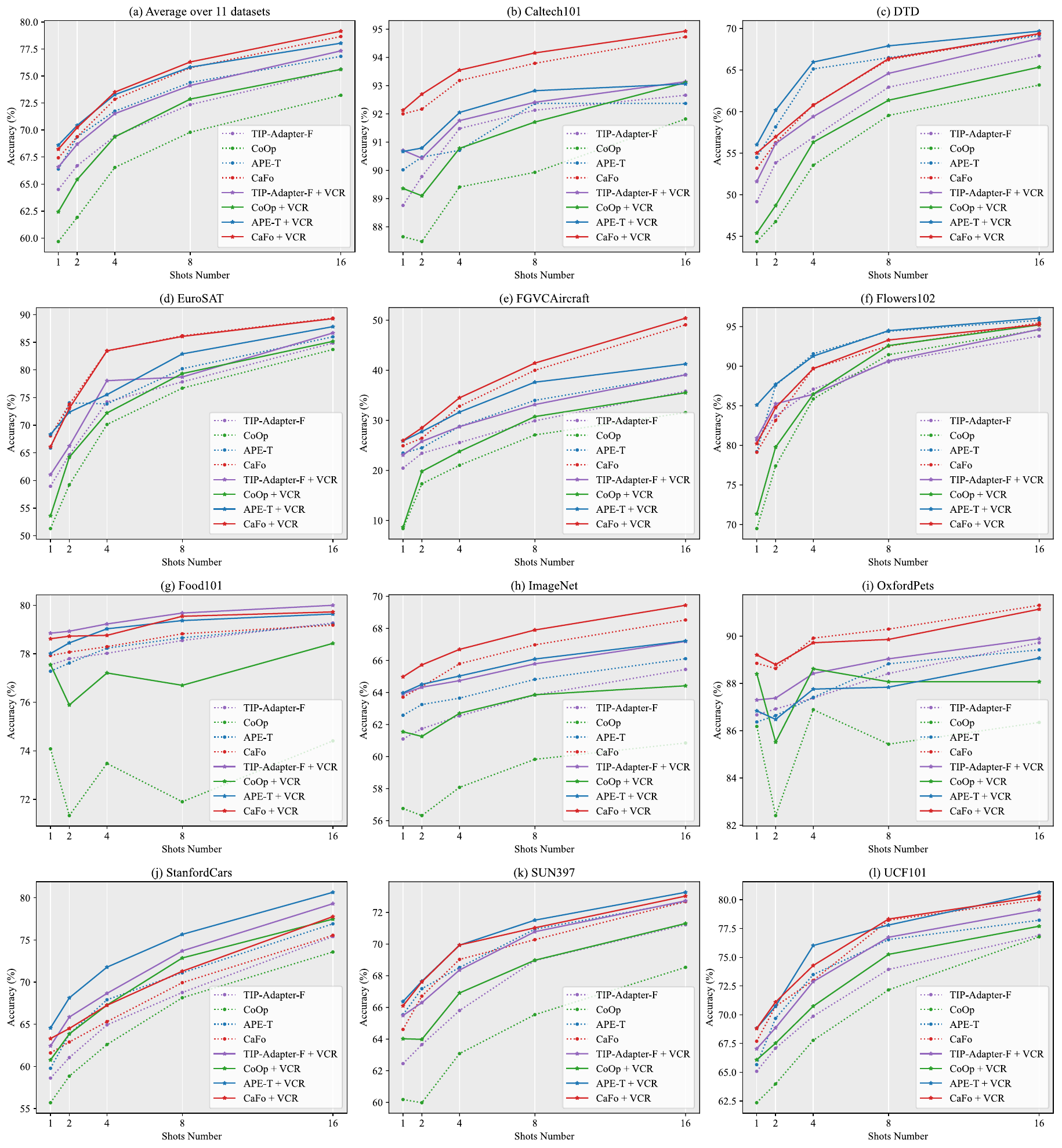}
\caption{Few-shot performance with training-need methods on 11 datasets, we first show the average results, and the following are organized in the order of dataset names, zoom in for clear recognition.}
\label{fig: training-need}
\end{figure*}

\subsection{Comparison with other methods.}
In this subsection, we compare our method with state-of-the-art methods under 3 different low-shot benchmark tasks, following are detailed illustrations of the experimental results.

\begin{table}[t]
\caption{Domain Generalization Performance (\%). The context length is 16 for CoOp \cite{zhou2022learning} with random initialization, and the training data consists of 16-shot ImageNet \cite{deng2009imagenet}.}
\centering
\begin{tabular}{l | c | c  c}
\hline\hline
\multirow{2}*{\textbf{Methods}} & \textbf{Source} & \multicolumn{2}{c}{\textbf{Target}}\\
 & ImageNet & -V2 & -Sketch \\ 
\hline
\multicolumn{4}{l}{\textbf{Training-free}} \\
\hline
Tip-Adapter & 62.00 & 54.60 & 35.93 \\
\textbf{ + VCR} & \textbf{64.49} & \textbf{57.28} & \textbf{37.67} \\
\hline
APE & 63.42 & 56.17 & 36.85 \\
\textbf{ + VCR} & \textbf{65.12} & \textbf{57.93} & \textbf{37.97} \\
\hline\hline
\multicolumn{4}{l}{\textbf{Training-need}} \\
\hline
Tip-Adapter-F & 65.44 & 57.17 & 36.05 \\
\textbf{ + VCR} & \textbf{67.19} & \textbf{59.11} & \textbf{37.61} \\
\hline
APE-T & 66.11 & 57.83 & 36.17 \\
\textbf{ + VCR} & \textbf{66.22} & \textbf{59.36} & \textbf{37.27} \\
\hline
CoOp & 60.85 & 53.09 & 32.57 \\
\textbf{ + VCR} & \textbf{64.42} & \textbf{56.92} & \textbf{35.73} \\
\hline\hline
\end{tabular}
\label{tab:domain-generalization}
\end{table}

\begin{table*}[t]
     \caption{Base-to-new generalization performance (\%). The context length is 4 with ``A photo of a'' initialization for all compared methods, and training data consists of 16-shot samples from each dataset. H is the harmonic mean \cite{zhou2022conditional}.}
    \begin{subtable}[t]{0.48\textwidth}
        \centering
        \caption{Average}
        \scalebox{0.9}{
        \begin{tabular}{l c c | c}
        \hline\hline
        Methods & Base & New & H \\
        \hline
        CoOp & 82.69 & 63.22 & 71.66\\
        \textbf{CoOp+VCR} & \textbf{84.28} & \textbf{66.37} & \textbf{73.40}\\
        \rowcolor{lightgray}
        \textbf{Improve  $(\uparrow)$} & \textcolor{purple}{+ 1.59} & \textcolor{purple}{+ 3.15} & \textcolor{purple}{+ 1.74} \\
        \hline\hline
       \end{tabular}
       }
       \label{tab:average}
    \end{subtable}
     \hfill
    \begin{subtable}[t]{0.48\textwidth}
        \centering
        \caption{ImageNet}
        \scalebox{0.9}{
        \begin{tabular}{l c c | c}
        \hline\hline
        Methods & Base & New & H \\
        \hline
        CoOp & 76.47 & 67.88 & 71.92\\
        \textbf{CoOp+VCR} & \textbf{77.45} & \textbf{70.58} & \textbf{73.86}\\
        \rowcolor{lightgray}
        \textbf{Improve  $(\uparrow)$} & \textcolor{purple}{+ 0.98} & \textcolor{purple}{+ 2.70} & \textcolor{purple}{+ 1.94} \\
        \hline\hline
       \end{tabular}
       }
       \label{tab:ImageNet}
    \end{subtable}
     \label{tab:base-to-new}
\end{table*}
\subsubsection{Domain Generalization.}
\label{subsubsec:exp_domain-generalization}

In this experiment, we evaluate our method under the domain generalization benchmark task. Specifically, we apply our method to Tip-Adapter \cite{zhang2022tip} and APE \cite{zhu2023not} for training-free comparison, and Tip-Adapter-F \cite{zhang2022tip}, APE-T \cite{zhu2023not}, CoOp \cite{zhou2022learning} for training-need comparison. All results are reported by our re-implementation with ResNet50 \cite{he2016deep} and the modified transformer \cite{radford2019language} as visual and textual encoders, respectively.
Following these methods, we utilize 16 training samples for each class from ImageNet, and then evaluate the classification accuracy on ImageNet-V2 and ImageNet-Sketch datasets. The experimental results are listed in Table~\ref{tab:domain-generalization}, where ``VCR'' denotes our Visual Content Refinement.

Specifically, we can find that our VCR not only improves the in-domain performance but also boosts the domain generalization performance. Specifically, VCR achieves more than 1.5\% and 1\% improvement over all compared methods on ImageNet-V2 and ImageNet-Sketch, respectively.
Meanwhile, the most improvement is achieved on CoOp with more than 3\% performance improvement on both datasets. 
Moreover, the proposed ``APE-T + VCR'' and ``APE + VCR'' attain the best performance on ImageNet-V2 and ImageNet-Sketch with 59.36\% and 37.97\% accuracy, respectively. 
Both the improvements and the attained performances are significant. 

\subsubsection{Few-shot Classification.}
\label{subsec:exp_few-shot}
We follow existing methods \cite{zhou2022learning,zhang2022tip,udandarao2023sus} to conduct few-shot classification with 1/2/4/8/16-shot training samples for both training-free and training-need settings. All results are reported by our re-implementation with ResNet50 \cite{he2016deep} and the modified transformer \cite{radford2019language} as visual and textual encoders, respectively.

\noindent \textbf{Training-free settings.} We apply our method on Tip-Adapter \cite{zhang2022tip}, Tip-X \cite{udandarao2023sus} and APE \cite{zhu2023not} for training-free comparison, the experimental results are shown in Figure~\ref{fig: training-free}, where ``VCR'' denotes our Visual Content Refinement.
For convenient illustration, the compared methods are in the same color as the respective applied ones, all compared methods are shown in dashed lines, and the applied methods are illustrated in solid lines. 

From the average results, we can find that our VCR yields remarkable performance enhancement over all compared methods, which achieves more than 1\% accuracy improvements across all shot settings, especially for Tip-Adapter, it brings 2.5\% and 2\% performance improvement for 1-shot and 16-shot settings, respectively. 
Meanwhile, we can observe that our VCR brings a significant improvement to the compared methods on StanfordCars. Specifically, it achieves more than 2\% improvement on Tip-X for all shot settings and more than 2.5\% improvement on Tip-Adapter and APE for all shot settings. 
Moreover, ``APE + VCR'' achieves new state-of-the-art for 16-shot setting, with 74.38\% and 65.12\% accuracy on StanfordCars and ImageNet, respectively.

\noindent \textbf{Training-need settings.} For training-need comparison, we utilize our VCR on Tip-Adapter \cite{zhang2022tip}, APE \cite{zhu2023not}, CaFo \cite{zhang2023prompt}, and CoOp \cite{zhou2022learning}. Figure~\ref{fig: training-need} shows the experimental results, where dashed lines represent the compared methods, solid lines correspond to our methods, and our applied methods are in the same color as the compared one, our method is described as ``VCR''. 

The average results over 11 datasets demonstrate that our VCR achieves significant improvement over the compared methods, especially for CoOp, which showcases more than 2.4\% performance improvement across all shot settings. 
Meanwhile, our VCR improves significantly for all compared methods on StanfordCars, it achieves more than 1.3\% improvement for CaFo and more than 3.5\% for other methods.
Furthermore, for the 16-shot setting, the proposed ``APE + VCR'' attains 80.66\% accuracy on StanfordCars, and the ``CaFo + VCR'' obtains 69.44\% performance on ImageNet, respectively, both achieving new state-of-the-art.

\subsubsection{Base-to-New Generalization.}
\label{subsubsec:exp_base-to-new}

In this experiment, we evaluate our VCR under the base-to-new generalization benchmark task. Specifically, we follow the work in CoOp \cite{zhou2022learning}, which trains the model on base categories, and evaluates the model on both base and new categories. 
The experimental results are shown in Table~\ref{tab:base-to-new}, where the experimental results are reported by following \cite{zhou2022conditional} with ViT-B/16 \cite{dosovitskiy2021an} and the modified transformer \cite{radford2019language} as visual and textual encoders, respectively. For simplicity, we list the average results over 11 datasets in Table~\ref{tab:average} and results over ImageNet dataset in Table~\ref{tab:ImageNet}.

Specifically, from Table~\ref{tab:average} we find that VCR brings a significant improvement over CoOp, 
Notably, the proposed ``CoOp + VCR'' outperforms CoOp with an average of 1.59\% and 3.15\% performance gains on base and new categories, respectively. 
Furthermore, on the challenging ImageNet dataset (Table~\ref{tab:ImageNet}), our VCR boosts the CoOp method with approximately 1\% and 2.7\% performance gains on both base and new categories, which is significant.

\begin{figure*}[t]
\centering
\includegraphics[width=1\textwidth]{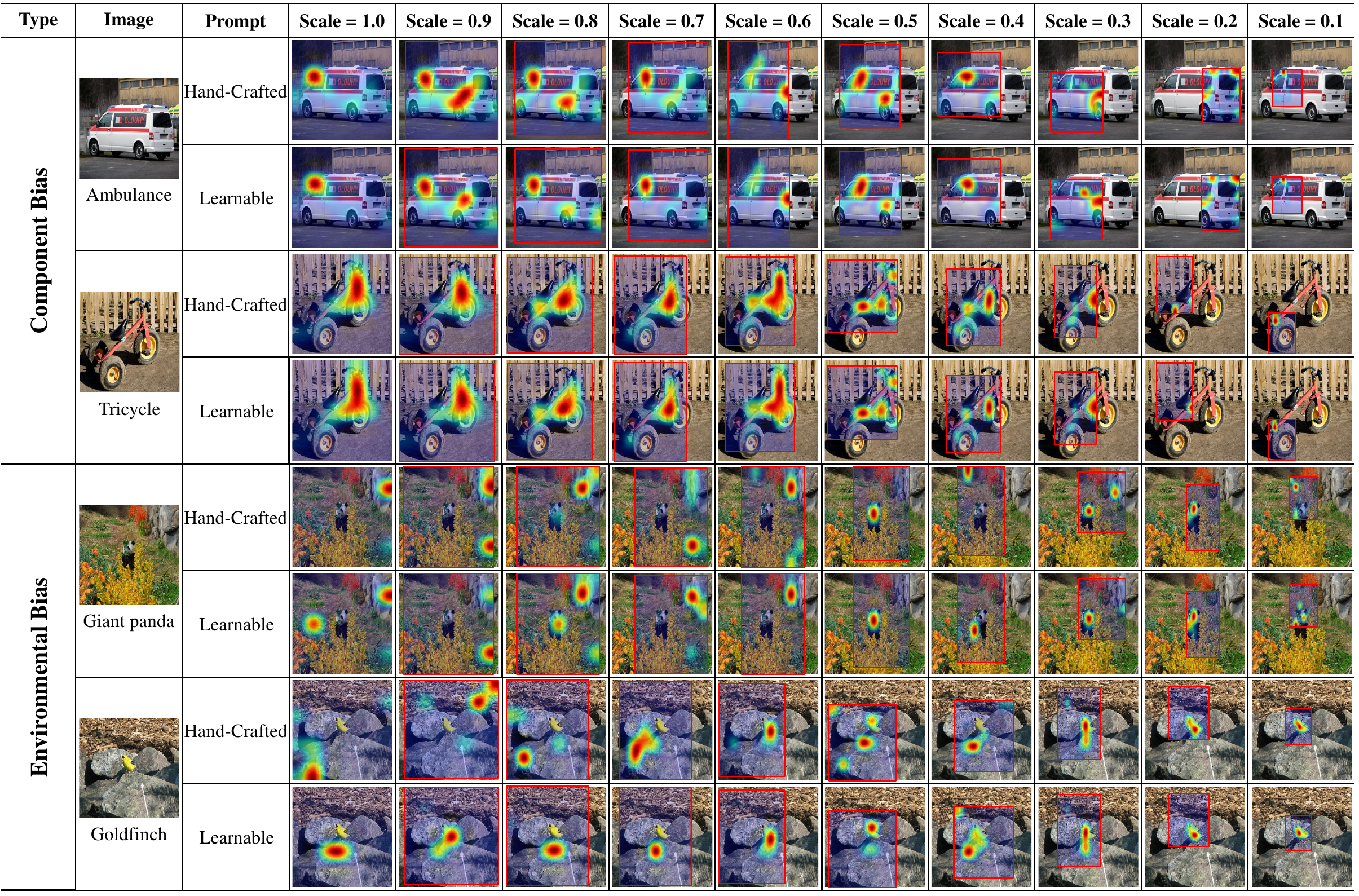}
\caption{The selected image regions from different scales by our VCR, where the responsive regions of the pre-trained CLIP model \cite{radford2021learning} are visualized by Grad-CAM \cite{selvaraju2017grad} from samples in the validation set of ImageNet \cite{deng2009imagenet}, ``Hand-Crafted \cite{zhang2022tip}'' and ``Learnable \cite{zhou2022learning}'' are two different prompt strategies, and ``Component Bias'' and ``Environmental Bias'' are two perceived bias issues.}
\label{fig:visualization}
\end{figure*}

\subsection{Computational Costs}
\label{subsec:computation}

\begin{table}[t]
\caption{Computational costs of VCR, all results are calculated under a single RTX 2080Ti GPU.}
\centering
\begin{tabular}{l | c c c}
\hline\hline
\textbf{Datasets} & \textbf{Times} & \textbf{Memory} & \textbf{Data Size} \\ 
\hline
\textbf{Caltech101} & 0.44s & 1288MB & 2465 \\
\textbf{FGVCAircraft} & 0.56s & 3540MB & 3333 \\
\textbf{Flowers102} & 0.46s & 1642MB & 2463 \\
\textbf{OxfordPets} & 0.66s & 1756MB & 3669 \\
\textbf{StanfordCars} & 1.30s & 3818MB & 8041 \\
\textbf{UCF101} & 0.42s & 1868MB & 3783 \\
\hline\hline
\end{tabular}
\label{tab:computation}
\end{table}

In this subsection, we evaluate the efficiency of our method. Specifically, we analyze the time and memory costs over 6 datasets with one RTX 2080Ti GPU, and the results are listed in Table~\ref{tab:computation}. It's noteworthy that our VCR is training-agnostic, which implies that our strategy can enhance the performance of current methods, regardless of the shot numbers and the training strategies.
For instance, compared to the efficient baseline method, \textit{i.e.} The training-free Tip-Adapter~\cite{zhang2022tip}, which takes 1.26s and 2.34s for 1-shot and 16-shot, respectively, with additional 0.44s and utilizing 1288MB GPU memory, we can improve the performance from 87.18\% to 89.57\% for 1-shot, and from 90.10\% to 92.06\% for 16-shot, which is significant for few-shot tasks.

\subsection{Visualization}
\label{subsec:visualization}

In this subsection, we visualize the selected responsive regions by our VCR with ``Hand-Crafted \cite{zhang2022tip}'' and ``Learnable \cite{zhou2022learning}'' prompts of pre-trained CLIP model \cite{radford2021learning} in Figure~\ref{fig:visualization}, where samples are from the validation set of ImageNet \cite{deng2009imagenet}. 

Specifically, we can find that: after decomposition and refinement, the adaptation model can 
\textbf{(1)} focus on the missing components of the objects by the global views, such as the back, the rear wheel of the ambulance in Figure~\ref{fig:visualization}(``Ambulance'', Scale= 0.2, 0.3), and the back wheel of the tricycle in Figure~\ref{fig:visualization}(``Tricycle'', Scale=0.1, 0.4).
and \textbf{(2)} discard the influences of environmental biases and select the object, for instance, the model mitigates the external noises (\textit{e.g.} the tree and the stone in Figure~\ref{fig:visualization}, respectively) and focuses on the actual category, such as the Giant Panda in Figure~\ref{fig:visualization}(``Giant Panda'', Scale=0.2, 0.5) and the Goldfinch in Figure~\ref{fig:visualization}( ``Goldfinch'', Scale=0.1, 0.2, 0.3 or 0.5).
These qualitative examples further demonstrate our hypothesis and illustrate the effectiveness of our VCR, which helps construct image features with high quality.

\section{Conclusion}
\label{sec:conclusion}

In this paper, we discuss the perceived bias issue of CLIP adaptation methods and propose a \textbf{V}isual \textbf{C}ontent \textbf{R}efinement (VCR) method to address this issue. Specifically, (1) The visual decomposition is proposed to construct multi-scale content to retain more local details. (2) The content refinement is designed to boost the representation of the input image by filtering out irrelevant noise and then combining the most confident image view at each scale. (3) Extensive experiments on 13 datasets demonstrate the effectiveness of our proposed method.

\balance
{
    \bibliography{sn-bibliography}
}

\end{document}